\documentclass[11pt,a4paper]{article}
\usepackage[hyperref]{acl2019}
\usepackage{times}
\usepackage{latexsym}

\usepackage{url}

\usepackage{tabularx} 
\usepackage{multirow}
\usepackage{booktabs} 

\usepackage{graphicx}  

\aclfinalcopy 

\title{Multi-passage BERT: A Globally Normalized BERT Model for Open-domain Question Answering}

\author{
Zhiguo Wang, Patrick Ng, Xiaofei Ma, Ramesh Nallapati, Bing Xiang \\
AWS AI Labs\\
  \texttt{\{zhiguow, patricng, xiaofeim, rnallapa, bxiang\}@amazon.com} \\
  }

\date{}

\begin{document}
\maketitle
\begin{abstract}
BERT model has been successfully applied to open-domain QA tasks.
However, previous work trains BERT by viewing passages corresponding to the same question as independent training instances,
which may cause incomparable scores for answers from different passages.
To tackle this issue, we propose a multi-passage BERT model to globally normalize answer scores across all passages of the same question,
and this change enables our QA model find better answers by utilizing more passages.
In addition, we find that splitting articles into passages with the length of 100 words by sliding window improves performance by 4\%.
By leveraging a passage ranker to select high-quality passages, multi-passage BERT gains additional 2\%.
Experiments on four standard benchmarks showed that our multi-passage BERT outperforms all state-of-the-art models on all benchmarks.
In particular, on the OpenSQuAD dataset, our model gains 21.4\% EM and 21.5\% $F_1$ over all non-BERT models, and 5.8\% EM and 6.5\% $F_1$ over BERT-based models. \footnote{To appear in EMNLP 2019.}
\end{abstract}

\section{Introduction}
BERT model \cite{devlin2018bert} has achieved significant improvements on a variety of NLP tasks.
For question answering (QA), it has dominated the leaderboards of several machine reading comprehension (RC) datasets.
However, the RC task is only a simplified version of the QA task, where a model only needs to find an answer from a given passage/paragraph.
Whereas, in reality, an open-domain QA system is required to pinpoint answers from a massive article collection, such as Wikipedia or the entire web.

Recent studies directly applied the BERT-RC model to open-domain QA \cite{yang2019end,nogueira2018learning,alberti2019bert}.
They firstly leverage a passage retriever to retrieve multiple passages for each question.
During training, passages corresponding to the same question are taken as independent training instances.
During inference, the BERT-RC model is applied to each passage individually to predict an answer span, 
and then the highest scoring span is selected as the final answer.
Although this method achieves significant improvements on several datasets, there are still several unaddressed issues.
\textbf{First}, viewing passages of the same question as independent training instances may result in incomparable answer scores across passages. Thus, globally normalizing scores over all passages of the same question \cite{clark2017simple} may be helpful.
\textbf{Second}, previous work defines passages as articles, paragraphs, or sentences. However, \emph{the question of proper granularity of passages} is still underexplored.
\textbf{Third}, passage ranker for selecting high-quality passages has been shown to be very useful in previous open-domain QA systems \cite{wang2018r,lin2018denoising,pang2019has}. However, we do not know whether it is still required for BERT.
\textbf{Fourth}, most effective QA and RC models highly rely on explicit inter-sentence matching between questions and passages \cite{wang2016machine,wang2016multi,seo2016bidirectional,wang2017gated},
whereas BERT only applies self-attention layers over the concatenation of a question-passage pair.
It is unclear whether the inter-sentence matching still matters for BERT.

To answer these questions, we conduct a series of empirical studies on the OpenSQuAD dataset \cite{rajpurkar2016squad,wang2018r}.
Experimental results show that: 
(1) global normalization makes QA model more stable while pinpointing answers from large number of passages; 
(2) splitting articles into passages with the length of 100 words by sliding window brings 4\% improvements;
(3) leveraging a BERT-based passage ranker gives us extra 2\% improvements;
and (4) explicit inter-sentence matching is not helpful for BERT.
We also compared our model with state-of-the-art models on four standard benchmarks,
and our model outperforms all state-of-the-art models on all benchmarks.

\section{Model}
Open-domain QA systems aim to find an answer for a given question from a massive article collection.
Usually, a retriever is leveraged to retrieve $m$ passages $P=[P_1, ...,P_i, ...,P_m]$ for a given question $Q=(q^1, ...,q^{|Q|})$, 
where $P_i=(p_i^1, ...,p_i^{|p_i|})$ is the $i$-th passage, 
and $q^k\in Q$ and $p_i^j\in P_i$ are corresponding words.
A QA model will compute a score $Pr(a|Q,P)$ for each possible answer span $a$.
We further decompose the answer span prediction into predicting the start and end positions of the answer span $Pr(a|Q,P)=P_s(a_s|Q,P)P_e(a_e|Q,P)$, where $P_s(a_s|Q,P)$ and $P_e(a_e|Q,P)$ are the probabilities of $a_s$ and $a_e$ to be the start and end positions.

\textbf{BERT-RC} model assumes passages in $P$ are independent of each other.
The model concatenates the question $Q$ and each passage $P_i$ into a new sequence ``[CLS] $p_i^1,  ..., p_i^{|p_i|}$ [SEP] $q^1,  ..., q^{|Q|}$ [SEP]'', and applies BERT to encode this sequence.
Then the vector representation of each word position from BERT encoder is fed into two separate dense layers to predict the probabilities $P_s$ and $P_e$ \cite{devlin2018bert}.
During training, the log-likelihood of the correct start and end positions for each passage is optimized independently.
For passages without any correct answers, we set the start and end positions to be 0, which is the position for the first token [CLS]. 
During inference, BERT-RC model is applied to each passage individually to predict an answer,
and then the highest scoring span is selected as the final answer.
If answers from different passages have the same string, they are merged by summing up their scores.

\textbf{Multi-passage BERT}:  BERT-RC model normalizes probability distributions $P_s$ and $P_e$ for each passage independently, which may cause incomparable answer scores across passages.
To tackle this issue, we leverage the global normalization method  \cite{clark2017simple} to normalize answer scores among multiple passages, and dub this model as multi-passage BERT.
Concretely, all passages of the same question are processed independently as we do in BERT-RC until the normalization step.
Then, $softmax$ is applied to normalize all word positions from all passages.

\textbf{Passage ranker} reranks all retrieved passages,
and selects a list of high-quality passages for the multi-passage BERT model. 
We implement the passage ranker as another BERT model, which is similar to multi-passage BERT except that at the output layer it only predicts a single score for each passage based on the vector representation of the first token [CLS].
We also apply $softmax$ over all passage scores corresponding to the same question,
and train to maximize the log-likelihood of passages containing the correct answers.
Denote the passage score as $Pr(P_i|Q,P)$, then the score of an answer span from passage $P_i$ will be $Pr(P_i|Q,P)P_s(a_s|Q,P)P_e(a_e|Q,P)$.

\section{Experiments}
\textbf{Datasets}: We experiment on four open-domain QA datasets.
(1) \emph{OpenSQuAD}: question-answer pairs are from  SQuAD 1.1 \cite{rajpurkar2016squad}, but a QA model will find answers from the entire Wikipedia rather than the given context.
Following \citet{chen2017reading}, we use the 2016-12-21 English Wikipedia dump.
5,000 QA pairs are randomly selected from the original training set as our validation set, and the remaining QA pairs are taken as our new training set.
The original development set is used as our test set.
(2) \emph{TriviaQA}: TriviaQA unfiltered version \cite{joshi2017triviaqa} are used. 
Following \citet{pang2019has}, we randomly hold out  5,000 QA pairs from the original training set as our validation set, and take the remaining pairs as our new training set.
The original development set is used as our test set.
(3) \emph{Quasar-T} \cite{dhingra2017quasar} and (4) \emph{SearchQA} \cite{dunn2017searchqa} are leveraged with the official split. 

\textbf{Basic Settings}:
If not specified, the pre-trained BERT-base model with default hyper-parameters is leveraged.
ElasticSearch with BM25 algorithm is employed as our retriever for OpenSQuAD.
Passages for other datasets are from the corresponding releases.
During training, we use top-10 passages for each question plus all passages (within the top-100 list) containing correct answers.
During inference, we use top-30 passages for each question.
Exact Match (EM) and $F_1$ scores \cite{rajpurkar2016squad} are utilized as the evaluation metrics.
\subsection{Model Analysis}
\begin{table}[tbp]
	\small
	\begin{center}
		\begin{tabular}{clcc}
			\toprule
			No. & Model & EM & $F_1$\\
			\midrule

			1 & Single-sentence & 34.8 & 44.4\\
			2 & Length-50 & 35.5 & 45.2\\
			3 & Length-100 & 35.7 & 45.7\\
			4 & Length-200 & 34.8 & 44.7\\
			\midrule
			5 & w/o sliding-window (same as (3))	 & 35.7& 45.7\\
			6 & w/ sliding-window & 40.4 & 49.8\\
			\midrule
			7 & w/o passage ranker (same as (6)) & 40.4 & 49.8\\
			8 & w/ passage ranker & 41.3 & 51.7\\
			9 & w/ passage scores & \textbf{42.8} & \textbf{53.4}\\
			\midrule
			10 & BERT+QANet & 18.3 & 27.8\\
			11 & BERT+QANet (fix BERT) & 35.5 & 45.9\\
			12 & BERT+QANet (init. from (11)) & 36.2 & 46.4\\
			\bottomrule
		\end{tabular}
	\end{center}
	\vspace{-0.5em}
	\caption{Results on the validation set of OpenSQuAD.}
	\label{tab:open_squad_val}
	\vspace{-1.0em}
\end{table}

To answer questions from section 1, we conduct a series of experiments on OpenSQuAD dataset, and report the validation set results in Table \ref{tab:open_squad_val}.
Multi-passage BERT model is used for experiments.

\textbf{Effect of passage granularity}:
Previous work usually defines passages as articles \cite{chen2017reading}, paragraphs \cite{yang2019end}, or sentences \cite{wang2018r, lin2018denoising}.
We explore the effect of passage granularity regarding to the passage length, i.e., the number of words in each passage.
Each article is split into non-overlapping passages based on a fixed length.
We vary passage length among \{50, 100, 200\}, and list the results as models (2) (3) (4) in Table \ref{tab:open_squad_val}, respectively.
Comparing to single-sentence passages (model (1)), leveraging fixed-length passages works better, and passages with 100 words works the best.
Hereafter, we set passage length as 100 words. 

\begin{figure}[tbp]
	\begin{center}
		\includegraphics[width=0.45\textwidth]{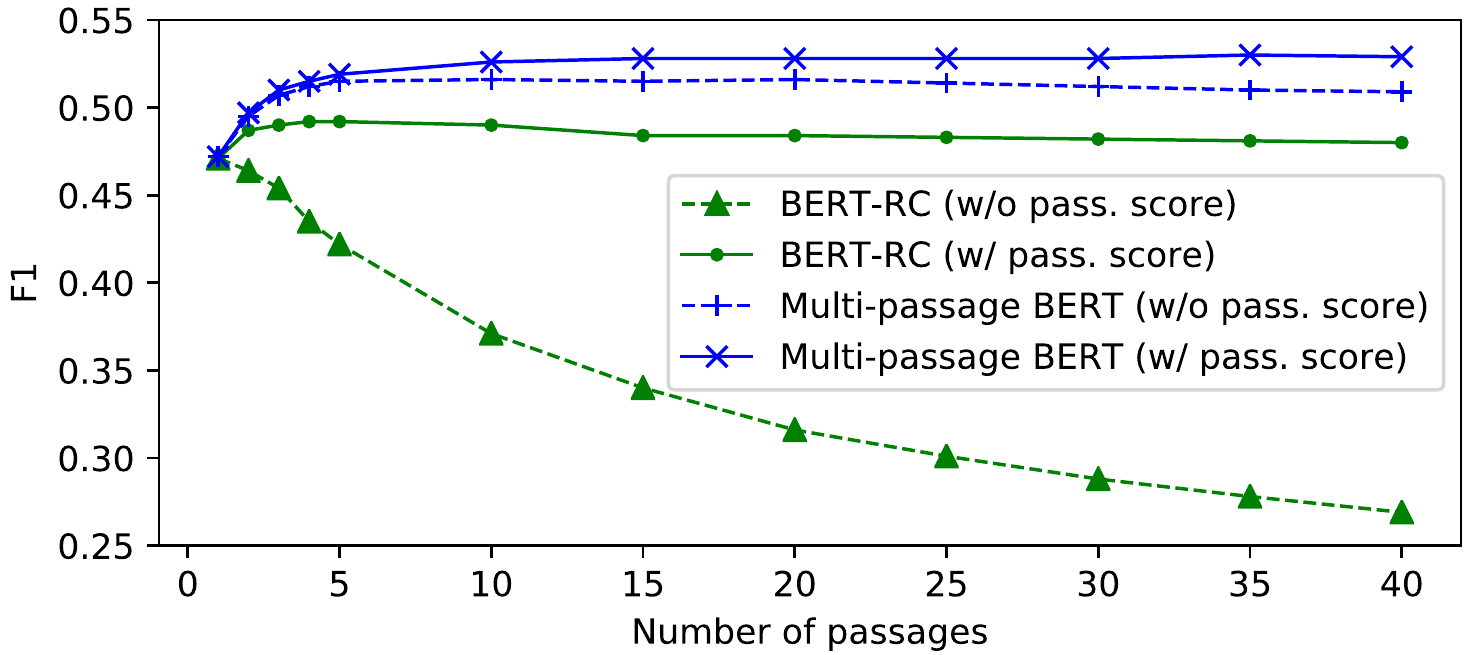}
	\end{center}
	\vspace{-0.5em}
	\caption{Effect of global normalization.}
	\label{fig:kbest_curve}
	\vspace{-1.0em}
\end{figure}

\textbf{Effect of sliding window}:
Splitting articles into non-overlapping passages may force some near-boundary answer spans to lose useful contexts.
To deal with this issue, we split articles into overlapping passages by sliding window.
We set the window size as 100 words, and the stride as 50 words (half the window size).
Result from the sliding window model is shown as model (6) in Table \ref{tab:open_squad_val}.
We can see that this method brings us 4.7\% EM and 4.1\% $F_1$ improvements.
Hereafter, we use sliding window method.

\begin{table*}[tbp]
	\small
	\begin{center}
		\begin{tabular}{lcccccccc}
			\toprule
			Datasets & \multicolumn{2}{c}{Quasar-T} & \multicolumn{2}{c}{SearchQA} & \multicolumn{2}{c}{TriviaQA} & \multicolumn{2}{c}{OpenSQuAD} \\
			\midrule
			Models & EM & $F_1$ & EM & $F_1$  & EM & $F_1$  & EM & $F_1$\\
			\midrule
			DrQA \cite{chen2017reading} & 37.7 & 44.5 & 41.9 & 48.7 & 32.3 & 38.3 & 29.8 & -\\
			$R^3$ \cite{wang2018r} & 35.3 & 41.7 & 49.0 & 55.3 & 47.3 & 53.7 & 29.1 & 37.5\\
			OpenQA \cite{lin2018denoising} & 42.2 & 49.3 & 58.8 & 64.5 & 48.7 & 56.3 & 28.7 & 36.6\\
			TraCRNet \cite{dehghani2019learning} & 43.2 & 54.0 & 52.9 & 65.1 & - &- &- &- \\
			HAS-QA \cite{pang2019has} & 43.2 & 48.9 & 62.7 & 68.7 & 63.6 & 68.9 & - & -\\

			\midrule
			BERT (Large) \cite{nogueira2018learning} & - & - & - & 69.1 & - & - & - & -\\
			BERT serini \cite{yang2019end} & - & - & - & - & - & - & 38.6 & 46.1\\
			BERT-RC (Ours) & 49.7 & 56.8 & 63.7 & 68.7 & 61.0 & 66.9 & 45.4 & 52.5 \\
			\midrule
			Multi-Passage BERT (Base) & \textbf{51.3} & 59.0 & \textbf{65.2} & 70.6 & 62.0 & 67.5 & 51.2 & 59.0\\
			Multi-Passage BERT (Large) & 51.1 & \textbf{59.1} & 65.1 & \textbf{70.7} & \textbf{63.7} & \textbf{69.2} & \textbf{53.0} & \textbf{60.9}\\
			\bottomrule
		\end{tabular}
	\end{center}
	\vspace{-0.5em}
	\caption{Comparison with state-of-the-art models, where the first group are models without using BERT, the second group are BERT-based models, and the last group are our multi-passage BERT models.}
	\label{tab:sota}
	\vspace{-1.0em}
\end{table*}

\textbf{Effect of passage ranker}:
We plug the passage ranker into the QA pipeline.
First, the retriever returns top-100 passages for each question. 
Then, the passage ranker is employed to rerank these 100 passages.
Finally, multi-passage BERT takes top-30 reranked passages as input to pinpoint the final answer.
We design two models to check the effect of the passage ranker.
The first model utilizes the reranked passages but without using passage scores, 
whereas the second model makes use of both the reranked passages and their scores.
Results are given in Table \ref{tab:open_squad_val} as models (8) and (9) respectively.
We can find that only using reranked passages gives us 0.9\% EM and 1.0\% $F_1$ improvements, 
and leveraging passage scores gives us 1.5\% EM and 1.7\% $F_1$ improvements.
Therefore, passage ranker is useful for multi-passage BERT model.

\textbf{Effect of global normalization}: 
We train BERT-RC and multi-passage BERT models using the reranked passages, 
then evaluate them by taking as input various number of passages.
These models are evaluated on two setups: with and without using passage scores.
$F_1$ scores for BERT-RC based on different number of passages are shown as the dotted and solid green curves in Figure \ref{fig:kbest_curve}.
$F_1$ scores for our multi-passage BERT model with similar settings are shown as the dotted and solid blue curves.
We can see that all models start from the same $F_1$, because multi-passage BERT is equivalent to BERT-RC when using only one passage.
While increasing the number of passages, BERT-RC without using passage scores decreases the performance significantly, which verifies that the answer scores from BERT-RC are incomparable across passages.
This issue is alleviated to some extent by leveraging passage scores.
On the other hand, performance from multi-passage BERT without using passage scores increases at the beginning, and then flattens out after passage number is over 10.
By utilizing passage scores, multi-passage BERT gets better performance while using more passages.
This phenomenon shows the effectiveness of global normalization, which enables the model find better answers by utilizing more passages.

\textbf{Does explicit inter-sentence matching matter?}
Almost all previous state-of-the-art QA and RC models find answers by matching passages with questions, aka inter-sentence matching \cite{wang2016machine,wang2016multi,seo2016bidirectional,wang2017gated,song2017unified}.
However, BERT model simply concatenates a passage with a question, and differentiates them by separating them with a delimiter token [SEP], and assigning different segment ids for them.
Here, we aim to check whether explicit inter-sentence matching still matters for BERT.
We employ a shared BERT model to encode a passage and a question individually, and a weighted sum of all BERT layers is used as the final token-level representation for the question or passage, where weights for all BERT layers are trainable parameters.
Then the passage and question representations are input into QANet \cite{yu2018qanet} to perform inter-sentence matching, and predict the final answer.
Model (10) in Table \ref{tab:open_squad_val} shows the result of jointly training the BERT encoder and the QANet model.
The result is very poor, likely because the parameters in BERT are catastrophically forgotten while training the QANet model.
To tackle this issue, we fix parameters in BERT, and only update parameters for QANet.
The result is listed as model (11).
It works better than model (10), but still worse than multi-passage BERT in model (6).
We design another model by starting from model (11), and then jointly fine-tuning the BERT encoder and QANet.
Model (12) in Table \ref{tab:open_squad_val} shows the result.
It works better than model (11), but still has a big gap with multi-passage BERT in model (6) .
Therefore, we conclude that the explicit inter-sentence matching is not helpful for multi-passage BERT.
One possible reason is that the multi-head self-attention layers in BERT has already embedded the inter-sentence matching.

\subsection{Comparison with State-of-the-art Models}

We evaluate BERT-RC and Multi-passage BERT on four standard benchmarks, where passage scores are leveraged for both models.
We build another multi-passage BERT for each dataset by initializing it with the pre-trained BERT-Large model.
Experimental results from our models as well as other state-of-the-art models are shown in Table \ref{tab:sota},
where the first group are open-domain QA models without using the BERT model, the second group are BERT-based models, and the last group are our multi-passage BERT models.

From Table \ref{tab:sota}, we can see that our multi-passage BERT model outperforms all state-of-the-art models across all benchmarks, 
and it works consistently better than our BERT-RC model which has the same settings except the global normalization.
In particular, on the OpenSQuAD dataset, our model improves by 21.4\% EM and 21.5\% $F_1$ over all non-BERT models, and 5.8\% EM and 6.5\% $F_1$ over BERT-based models \footnote{Our result for openSQuAD in the test set is significantly better than results in the dev set in Table \ref{tab:open_squad_val}, because our test set is from the official development set of SQuAD 1.1, where each question contains more than 3 annotated answers, whereas each question contains only one gold-standard answer in our dev set.}.
Leveraging BERT-Large model makes multi-passage BERT even better on TriviaQA and OpenSQuAD datasets.

\section{Conclusion}
We propose a multi-passage BERT model for open-domain QA to globally normalize answer scores across mutiple passages corresponding to the same question.
We find two effective techniques to improve the performance of multi-passage BERT: 
(1) splitting articles into passages with the length of 100 words by sliding window;
and (2) leveraging a passage ranker to select high-quality passages.
With all these techniques, our multi-passage BERT model outperforms all state-of-the-art models on four standard benchmarks.

In future, we plan to consider inter-correlation among passages for open-domain question answering \cite{wang2017evidence,song2018exploring}.


\bibliography{acl2019}
\bibliographystyle{acl_natbib}

\end{document}